\documentclass[letterpaper]{article} % DO NOT CHANGE THIS
\usepackage[preprint]{aaai2027}
\usepackage[hyphens]{url}
\usepackage{natbib}
\usepackage{caption}
\usepackage[utf8]{inputenc}
\usepackage[T1]{fontenc}
\usepackage{graphicx}
\usepackage{xcolor}
\usepackage{booktabs}
\usepackage{amsmath,amssymb,amsfonts,amsthm}
\usepackage{microtype}
\usepackage{bm}
\usepackage{subcaption}
\usepackage{multirow}

\newtheorem{definition}{Definition}

\newcommand{\supertype}[1]{g(#1)}
\newcommand{\supertypefunc}{g}

\newcounter{appsec}\renewcommand{\theappsec}{\Alph{appsec}}
\newcommand{\appsec}[2]{\refstepcounter{appsec}\label{#1}\paragraph{\theappsec. #2}}

\begin{document}

\title{Subtype Robustness Is Not Just Accuracy:\\
Calibration Under Unseen Subtype Shift}
\author{
    Hanyu Su\textsuperscript{1},
    Carlota Julbe i Juanola\textsuperscript{1},
    Yibo Hu\textsuperscript{1}
}
\affiliations{
    \textsuperscript{1}Illinois Institute of Technology\\
    \texttt{\{hsu18, cjulbeijuanola\}@hawk.illinoistech.edu, yhu89@illinoistech.edu}
}
\maketitle

%======================= ABSTRACT =======================
\begin{abstract}
Subtype robustness asks whether a model keeps the correct coarse prediction when
test examples come from fine-grained subtypes absent from training but still
inside a known coarse category. Prior work studies this almost entirely through
accuracy. We ask whether the model also stays calibrated. We present the first
systematic study of the question across ImageNet, BREEDS, iNaturalist and
CIFAR-100 with five architectures. Calibration breaks down on
unseen subtypes, where accuracy drops while confidence barely follows, leaving
the model systematically overconfident exactly where it has become less accurate.
At matched accuracy loss, generic image corruption causes a much larger drop in confidence, so
the effect is not a general consequence of losing accuracy. The model reacts to
visible degradation but not to in-taxonomy novelty. Recalibration tuned on seen subtypes narrows the gap but
does not close it, and out-of-distribution scores flag the affected inputs only
weakly. Subtype robustness should therefore be evaluated through calibration, not
accuracy alone.\footnote{The code and data are available at\\
\url{https://github.com/yibo-hu-lab/subtype-robustness}}
\end{abstract}

%======================= 1. INTRODUCTION =======================
\section{Introduction}

\begin{figure*}[t]
  \centering  \includegraphics[width=\textwidth]{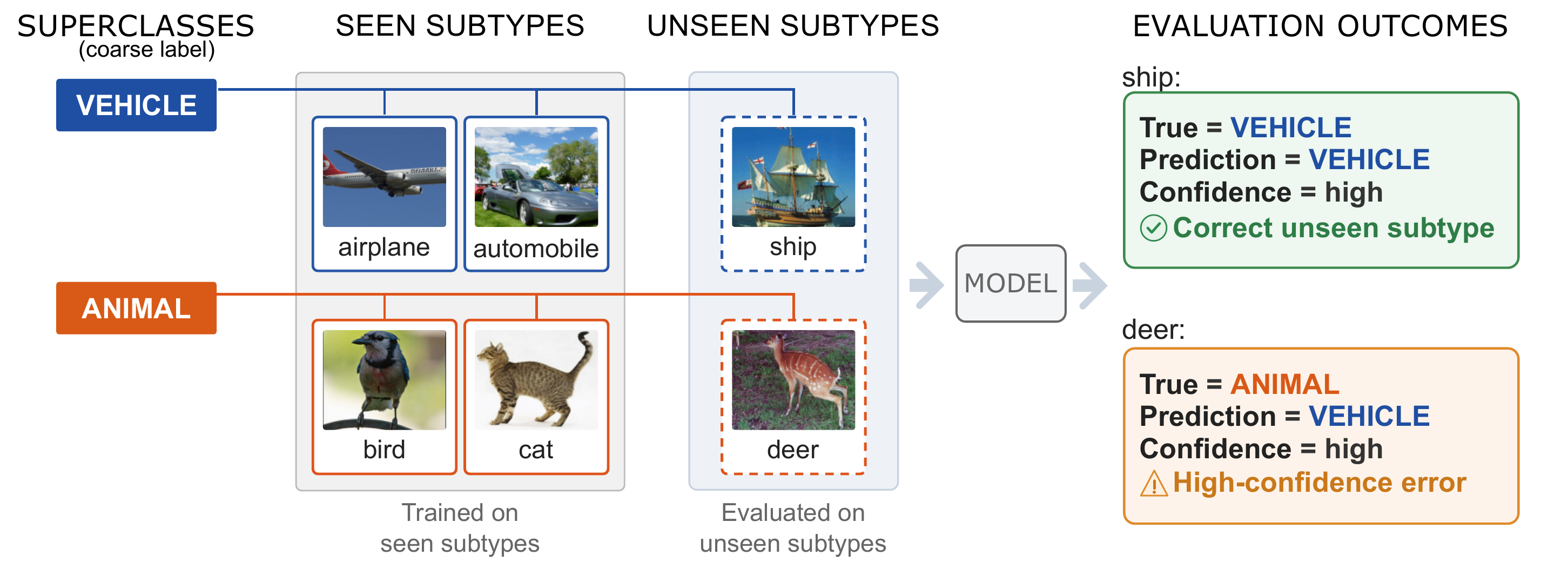}
        \caption{Toy example of subtype robustness. The model is trained on \textit{airplane} and \textit{automobile} (\textit{vehicle}), and \textit{bird} and \textit{cat} (\textit{animal}), then evaluated on unseen subtypes \textit{ship} and \textit{deer} from the same supertypes. The unseen \textit{ship} is correctly predicted as \textit{vehicle}, and the unseen \textit{deer} is misclassified as \textit{vehicle} instead of \textit{animal}, yet the model is equally confident in both.}\label{fig:toy_subtype}
\end{figure*}

Deep learning systems are often deployed in settings where the correct
semantic decision remains unchanged, even though its fine-grained
realizations continue to vary. In hierarchical label spaces, this means that
test examples may come from fine-grained subtypes absent from training, while
their correct coarse labels remain within the known taxonomy. This is the
setting of \textbf{subtype robustness}: the model need not recover the exact
unseen subtype, but it should still preserve the correct coarse semantic
prediction. Figure~\ref{fig:toy_subtype} illustrates it.

This setting is distinct from classical out-of-distribution (OOD) detection: the input still carries a valid coarse label, so the failure is a wrong coarse prediction. In many applications the coarse decision is the primary
target, and failing to recover the exact subtype costs less than predicting the
wrong coarse category~\cite{bertinetto2020making,garg2022learning}. Real-world
systems routinely meet new variants of known categories while the coarse decision
stays the same. A species classifier deployed for biodiversity monitoring, for
example, continually meets newly catalogued species under a known genus or family:
the fine-grained label is novel, the coarse taxonomic decision is not.

Prior work on subtype robustness has focused primarily on accuracy.
\citet{hendrycks2019benchmarking} reported seen--unseen subtype gaps, and
\citet{santurkar2021breeds} introduced BREEDS for controlled subtype-split
evaluation. Accuracy alone, however, gives an incomplete view. A model's coarse
accuracy can fall on unseen subtypes while its confidence barely follows, leaving
it systematically overconfident on the predictions it now gets wrong.
This overconfidence is a calibration failure: a calibrated model's
confidence matches its accuracy \citep{guo2017calibration}, so confidence that
stays high while accuracy falls is by definition miscalibrated. To our knowledge,
calibration under subtype-level novelty is unexplored.

Common image corruptions behave differently. As the image degrades, models become
less accurate and less confident, as observed broadly under distribution
shift~\citep{ovadia2019can}: the image looks different, and the model reacts.
Subtype shift breaks that link, because the image looks normal and its coarse
label is still correct.

We formalize subtype robustness as a same-coarse-label, unseen-fine-subtype
shift, read off a single model through matched accuracy, confidence, and
calibration gaps ($\Delta_{\mathrm{acc}}$, $\Delta_{\mathrm{conf}}$,
$\Delta_{\mathrm{cal}}$), and make three contributions.
\begin{itemize}
  \item \textbf{Confidence does not follow accuracy.} Across ImageNet, BREEDS
  (Living-17, Nonliving-26), iNaturalist, and CIFAR-100, and across
  architectures, coarse accuracy drops on unseen subtypes while confidence barely
  follows, leaving the model overconfident where it has become less accurate. The
  size of the accuracy drop varies by dataset: steep where the taxonomy is sparse,
  mild where it is densely sampled. The signal also fails per sample:
  failure-detection AUROC and risk--coverage both degrade.
  \item \textbf{The failure does not appear under matched generic corruption.} At
  matched accuracy loss, generic image corruption lowers confidence far more than
  subtype novelty does.
  \item \textbf{The post-hoc tools we test do not recover the lost calibration.} Recalibration
  tuned on seen subtypes narrows but does not close the seen--unseen calibration
  gap, and inference-time confidence and OOD scores flag unseen subtypes only
  weakly.
\end{itemize}

%======================= 2. RELATED WORK =======================
\section{Related Work}

\paragraph{Subtype robustness and subpopulation shift.}
The subtype robustness setting traces to \citet{hendrycks2019benchmarking}, whose
Appendix G measures accuracy on held-out subtypes of known coarse categories, and
it has remained largely under-explored since. The closest benchmark continuation
is BREEDS, which constructs hierarchy-aware source--target splits with disjoint
subtype support under curated ImageNet superclasses \cite{santurkar2021breeds}. In
the failure-detection literature, \citet{jaeger2023call} include a CIFAR-100
``sub-class shift'' condition in their evaluation of misclassification-detection
metrics. Broader subpopulation- and
domain-shift benchmarks provide an umbrella,
though the underlying shift definitions differ \cite{koh2021wilds,yang2023change}.

\paragraph{Hierarchical classification and semantic error structure.}
A related line studies hierarchy-aware prediction, where errors within a taxonomy
are not equally severe and a coarse-label error is treated differently from a
fine-grained one
\cite{deng2010does,barz2019hierarchy,bertinetto2020making,karthik2021no,garg2022learning,jain2023test}.

\paragraph{Calibration and recalibration under shift.}
Calibration methods assess whether model confidence is aligned with empirical
correctness~\cite{guo2017calibration,widmann2019calibration,minderer2021revisiting}.
Modern networks can become substantially miscalibrated under distribution shift,
including OOD and novel-class settings~\cite{hendrycks2018deep,kong2020calibrated,hu2021multidimensional,hu2021uncertainty}.
Large pretrained models are reported to be better calibrated and more robust to
such shift~\cite{minderer2021revisiting}. A common fix for calibration under shift is to
re-weight the calibration data toward the test
distribution~\cite{wang2025conformal,cofact2026,compass2026,moecp2026}. Finally,
\citet{lyu2025imbalance} offer a candidate mechanism: under-represented groups
receive inflated training logits. Our setting is the extreme case, because unseen
subtypes are absent from training rather than merely rare.

\paragraph{Out-of-distribution detection.}
Detection methods flag inputs whose label lies outside the taxonomy, using
output-based scores such as the maximum softmax
probability~\cite{hendrycks2017baseline} and the energy score~\cite{liu2020energy},
or feature-based scores such as the Mahalanobis distance~\cite{lee2018simple}. We
therefore ask whether these scores respond to a novel subtype whose coarse label
is still valid. Benchmarks in this area have been argued to conflate semantic with
covariate shift, with detectors responding mainly to the
former~\cite{zhu2024imagenetood}. Subtype shift fits neither cleanly: the label
stays inside the taxonomy, as in covariate shift, yet what changes is semantic
rather than appearance.

%======================= 3. PROBLEM SETUP =======================
\section{Problem Setup}

\subsection{Hierarchical classification}
Multiclass classification maps instances \(x \in \mathcal{X}\) to one of
\(K \ge 2\) discrete labels \(\mathcal{K} := \{1,\dots, K\}\). We assume these
labels are partitioned into \(J\) mutually exclusive, nonempty subsets
\(\mathcal{K}_{1},\dots,\mathcal{K}_{J}\) indexed by
\(\mathcal{J} := \{1,\dots, J\}\), with
\(2 \le J \le K\). We call the labels \(k \in \mathcal{K}\) \textbf{subtypes} and
the parts \(j \in \mathcal{J}\) \textbf{supertypes}, and write \(\supertype{k}\)
for the supertype of subtype \(k\), so that
\(\supertype{\cdot} : \mathcal{K} \to \mathcal{J}\) is surjective.

\begin{definition}[Seen/unseen subtypes]
We assume a nonempty subset \(\mathcal{S} \subseteq \mathcal{K}\) of
\textbf{seen subtypes}, whose complement
\(\mathcal{U} := \mathcal{K} \setminus \mathcal{S}\) is the set of
\textbf{unseen subtypes}, so that \(1 \le |\mathcal{S}| < K\). Every supertype is
represented among the seen subtypes:
\(\supertype{\mathcal{S}} = \supertype{\mathcal{U}} = \mathcal{J}\).
\end{definition}
Each part \(\mathcal{K}_{j} = \supertypefunc^{-1}(j)\) therefore splits into
\(\mathcal{S}_{j} := \mathcal{K}_{j} \cap \mathcal{S}\), the seen subtypes under
\(j\), and \(\mathcal{U}_{j} := \mathcal{K}_{j} \cap \mathcal{U}\), the unseen
ones. We write \(n_{j} := |\mathcal{S}_{j}|\) for how many seen subtypes a
supertype has. The Analysis section relates this quantity to the size of the
calibration collapse.

\subsection{Training scheme}
All models are trained using only examples whose subtype labels belong to
\(\mathcal{S}\), and predict the supertype. The key constraint is that no examples
from \(\mathcal{U}\) are used for representation learning, classifier training,
calibration, or model selection, except in the post-hoc calibration experiment where
we state otherwise. Model selection and any post-hoc
calibrator are fit on \emph{seen-val}, a held-out set of \emph{images} drawn from
the same seen subtypes used for training, never on test. Every split is carved at
the image level within \(\mathcal{S}\), so no subtype label is withheld from
training. The four-way
train\,/\,seen-val\,/\,test-seen\,/\,test-unseen split is detailed in Appendix~\ref{app:datasets}.

\paragraph{Training procedure.}
All main-table models are trained \emph{direct-coarse}: a single network optimized end-to-end on the supertype label from a random initialization, uniformly across all datasets, as in BREEDS~\cite{santurkar2021breeds}. \citet{hendrycks2019benchmarking} instead reuse a fine-pretrained frozen backbone, a two-stage protocol on a pretrained network. We report the protocol (one- vs.\ two-stage) and initialization (from-scratch vs.\ ImageNet-pretrained) axes as controls in Appendix~\ref{app:protocol}, where the protocol is neutral and pretraining mitigates the overconfidence without removing it.

\subsection{Evaluation scheme}
\paragraph{Accuracy.}
We evaluate on both seen and unseen subtype splits, always scoring the
predicted supertype. For an example \((x,y)\), the coarse-label error is
\(\mathbf{1}\{\widehat{j}(x) \ne \supertype{y}\}\). Accuracy is computed
separately on examples with \(y\in\mathcal{S}\) and \(y\in\mathcal{U}\).

\paragraph{Calibration.}
Let \(p_j(x)\) denote the model probability assigned to
supertype \(j\). We define the confidence and correctness of a prediction as
\begin{align*}
c(x) &= \max_{j\in\mathcal{J}} p_j(x), \\
r(x,y) &= \mathbf{1}\{\arg\max\nolimits_{j\in\mathcal{J}} p_j(x)=\supertype{y}\}.
\end{align*}
We measure calibration with the expected calibration error (ECE) over $15$
equal-width bins~\cite{naeini2015obtaining} (Appendix~\ref{app:binning} covers the
estimator and its binning bias).
Given that ECE is a binned estimator, we corroborate every gap with an equal-mass
recomputation and with two binning-free proper scores~\cite{gneiting2007proper}, the
negative log-likelihood (NLL) and the Brier score~\cite{brier1950}, all in
Appendix~\ref{app:proper}. Since ECE averages
$|\widehat{\mathrm{acc}}-\widehat{\mathrm{conf}}|$, all three quantities share one
scale, and we report accuracy, confidence, ECE, and the seen$-$unseen gaps defined
below in \textbf{percentage points} throughout.

\paragraph{Per-sample discrimination.}
Calibration is an aggregate property and can be orthogonal to whether confidence
discriminates correct from incorrect predictions per sample.On each split we therefore also report the two standard failure-detection
measures~\cite{jaeger2023call}: the areas under the ROC curve (AUROC, whether
\(c(x)\) separates correct from incorrect predictions) and the risk--coverage
curve (AURC, lower is better). Both are rank-based, so unlike
the calibration gap, a loss here cannot be repaired by rescaling confidence.

\paragraph{Subtype accuracy, confidence, and calibration gaps.}
We summarize each model by three matched seen-versus-unseen gaps,
$\Delta_{\mathrm{acc}}=\mathrm{acc}_{\mathcal{S}}-\mathrm{acc}_{\mathcal{U}}$,
$\Delta_{\mathrm{conf}}=\bar{c}_{\mathcal{S}}-\bar{c}_{\mathcal{U}}$, and
$\Delta_{\mathrm{cal}}=\mathrm{ECE}_{\mathcal{U}}-\mathrm{ECE}_{\mathcal{S}}$, where
\(\bar{c}_{\bullet}\) is the mean confidence \(c(x)\) on the split. A model is
\emph{calibration-robust} to subtype shift when \(\Delta_{\mathrm{cal}}\) stays
small. When instead \(\Delta_{\mathrm{conf}} \ll \Delta_{\mathrm{acc}}\), the two gaps
\emph{dissociate}, and we call the resulting failure \emph{silent overconfidence}.

%======================= 4. EXPERIMENTS =======================
\section{Experiments}

\subsection{Research Questions}
\textbf{RQ1 (Calibration under subtype shift).} When a model loses accuracy on
unseen subtypes, does it stay calibrated there, and does high confidence still
indicate a correct prediction?

\noindent\textbf{RQ2 (Comparison with generic corruption).} If generic image corruption is
tuned to cost the same accuracy as subtype shift, does it cost the same confidence?

\noindent\textbf{RQ3 (Standard remedies).} Can the calibration gap be closed after training,
either by (a) recalibrating on seen subtypes or (b) catching novel subtypes at
inference with standard OOD scores?

\subsection{Datasets and Evaluation Settings}
\textbf{ImageNet-25} (\textbf{IN-25})~\cite{deng2009imagenet} is the largest
of our benchmarks: 25 broad categories as supertypes, with
383 seen subtypes (ImageNet-1K synsets) and 3{,}872 held-out unseen subtypes
(ImageNet-22K-only synsets) under those same categories. Its number of seen
subtypes per supertype is deliberately uneven (\textit{dog} many, \textit{fungus}
few), inherited from ImageNet-1K, and we do not rebalance it.

\noindent\textbf{BREEDS}~\cite{santurkar2021breeds} is derived from the same ImageNet-1K
classes but purpose-built for hierarchy-aware seen/unseen subtype splits. We take
two tasks that span the living/man-made divide, \textbf{Living-17} (\textbf{Liv-17};
17 supertypes; natural, animate) and \textbf{Nonliving-26} (\textbf{NL-26}; 26
supertypes; man-made), each supertype split into two seen and two unseen subtypes.

\noindent\textbf{iNaturalist} (\textbf{iNat-25})~\cite{vanhorn2021inaturalist} carries a real Linnaean
taxonomy (order $\supseteq$ family $\supseteq$ genus $\supseteq$ species), so an
unseen subtype is literal: a held-out species under a known order. We use 25 orders
as supertypes, splitting each order's species into 3{,}427
seen and 3{,}427 held out. It is also the only benchmark that supplies a model-free
notion of subtype distance (taxonomic rank).

\noindent\textbf{CIFAR-100}~\cite{krizhevsky2009learning} is a low-resolution sanity check
rather than a primary benchmark: 20 supertypes, 60 of the 100 fine classes seen and 40 held out, at
$32\times32$ with two small models (AlexNet, ResNet-18).

\begin{table*}[t]
\centering
\small
\setlength{\tabcolsep}{6pt}
\caption{Accuracy, confidence and ECE on seen and unseen subtypes, with the
seen$-$unseen gap for each, all in \textbf{percentage points}.
On every row ECE is close to confidence minus accuracy, hence
$\Delta_{\mathrm{cal}} \approx \Delta_{\mathrm{acc}} - \Delta_{\mathrm{conf}}$.
Appendix~\ref{app:proper} checks the gap against equal-mass binning and against two binning-free
proper scores.}
\begin{tabular}{ll ccc ccc ccc}
\toprule
 & & \multicolumn{3}{c}{Accuracy $\uparrow$} & \multicolumn{3}{c}{Confidence} & \multicolumn{3}{c}{ECE $\downarrow$} \\
\cmidrule(lr){3-5}\cmidrule(lr){6-8}\cmidrule(lr){9-11}
\textbf{Dataset} & \textbf{Model} & Seen & Unseen & $\bm{\Delta_{\mathrm{acc}}}$ & Seen & Unseen & $\bm{\Delta_{\mathrm{conf}}}$ & Seen & Unseen & $\bm{\Delta_{\mathrm{cal}}}$ \\
\midrule
\multirow[c]{2}{*}{CIFAR-100}
  & AlexNet    & 72.0 & 37.2 & 34.8 & 89.3 & 82.0 & 7.2 & 17.3 & 44.9 & 27.6 \\
  & ResNet-18  & 84.7 & 44.8 & 39.9 & 90.9 & 78.5 & 12.4 & 6.2 & 33.8 & 27.5 \\
\midrule
\multirow[c]{5}{*}{IN-25}
  & AlexNet    & 80.7 & 51.3 & 29.4 & 82.8 & 72.7 & 10.1 & 2.1 & 21.4 & 19.3 \\
  & ResNet-18  & 90.0 & 60.0 & 30.0 & 92.6 & 81.2 & 11.3 & 2.6 & 21.2 & 18.6 \\
  & ResNet-50  & 91.0 & 60.7 & 30.3 & 93.4 & 82.3 & 11.2 & 2.5 & 21.7 & 19.2 \\
  & ViT-B/16   & 84.1 & 50.1 & 34.0 & 92.5 & 83.9 & 8.6 & 8.4 & 33.8 & 25.4 \\
  & ConvNeXt-T & 91.3 & 61.3 & 30.0 & 95.4 & 86.7 & 8.7 & 4.1 & 25.4 & 21.3 \\
\midrule
\multirow[c]{5}{*}{Liv-17}
  & AlexNet    & 86.8 & 53.8 & 33.0 & 91.2 & 81.4 & 9.8 & 5.2 & 27.7 & 22.5 \\
  & ResNet-18  & 90.1 & 61.4 & 28.7 & 94.0 & 86.0 & 8.0 & 4.0 & 25.0 & 21.0 \\
  & ResNet-50  & 89.2 & 56.8 & 32.4 & 93.4 & 84.4 & 8.9 & 4.5 & 27.6 & 23.1 \\
  & ViT-B/16   & 79.8 & 42.5 & 37.2 & 91.2 & 82.0 & 9.2 & 11.4 & 39.5 & 28.0 \\
  & ConvNeXt-T & 89.4 & 57.3 & 32.1 & 94.4 & 87.1 & 7.3 & 5.2 & 29.8 & 24.7 \\
\midrule
\multirow[c]{5}{*}{NL-26}
  & AlexNet    & 75.7 & 39.8 & 35.9 & 80.1 & 64.3 & 15.8 & 4.6 & 24.6 & 20.0 \\
  & ResNet-18  & 84.1 & 46.5 & 37.6 & 90.0 & 74.6 & 15.4 & 6.1 & 28.1 & 22.0 \\
  & ResNet-50  & 84.0 & 45.3 & 38.6 & 88.1 & 71.8 & 16.2 & 4.4 & 26.5 & 22.1 \\
  & ViT-B/16   & 63.2 & 30.3 & 32.9 & 84.4 & 76.0 & 8.5 & 21.2 & 45.8 & 24.5 \\
  & ConvNeXt-T & 80.9 & 42.9 & 38.0 & 89.8 & 76.6 & 13.2 & 8.9 & 33.7 & 24.8 \\
\midrule
\multirow[c]{5}{*}{iNat-25}
  & AlexNet    & 68.6 & 61.2 & 7.4 & 72.5 & 70.3 & 2.2 & 3.9 & 9.1 & 5.1 \\
  & ResNet-18  & 75.6 & 62.8 & 12.8 & 85.9 & 81.6 & 4.3 & 10.3 & 18.8 & 8.5 \\
  & ResNet-50  & 76.2 & 63.6 & 12.6 & 86.6 & 82.9 & 3.6 & 10.3 & 19.3 & 9.0 \\
  & ViT-B/16   & 65.4 & 56.2 & 9.2 & 87.2 & 85.2 & 2.0 & 21.8 & 29.0 & 7.2 \\
  & ConvNeXt-T & 79.5 & 69.6 & 9.8 & 91.7 & 89.6 & 2.2 & 12.3 & 19.9 & 7.7 \\
\bottomrule
\end{tabular}
\label{tab:gap-all-datasets}
\end{table*}

Across all datasets we evaluate on a per-supertype-balanced test set.
Appendix~\ref{app:datasets} gives full construction, splits, and per-class counts, and Appendix~\ref{app:construction}
details the iNaturalist taxonomy build.

\subsection{Models and Baselines}
We compare five architectures:
\begin{itemize}
  \item \textbf{Historical baseline:} AlexNet~\cite{krizhevsky2012alexnet}.
  \item \textbf{Residual models:} ResNet-18 and ResNet-50~\cite{he2016resnet}.
  \item \textbf{Modern backbones:} ViT-B/16~\cite{dosovitskiy2021vit} and ConvNeXt-T~\cite{liu2022convnext}.
\end{itemize}
On the full-resolution ($224\times224$) datasets (IN-25, BREEDS, iNat-25)
we sweep all five architectures; CIFAR-100 uses only the small models (AlexNet,
ResNet-18) that train competently at $32\times32$.

For calibration, we evaluate two standard post-hoc calibrators, tuned on a held-out
image split from the seen subtypes and then applied to unseen subtypes: temperature
scaling and vector scaling~\cite{guo2017calibration}.

For detection, we use three standard scores. Two read the output distribution:
maximum softmax probability (MSP)~\cite{hendrycks2017baseline} and the energy
score~\cite{liu2020energy}. The third reads the representation: the Mahalanobis
distance to the nearest seen-class feature centroid~\cite{lee2018simple}. An output
score can only flag what the model already doubts; a feature score can flag inputs
that look novel even when the output is confident.

The full training and evaluation recipe is given in
Table~\ref{tab:training-recipe}. The two controlled variations reported separately,
the fine-pretrained-frozen protocol on CIFAR-100 and BREEDS and an
ImageNet-1K-pretrained backbone on IN-25, are covered in
Appendix~\ref{app:protocol}.

\subsection{Accuracy drops but confidence barely follows (RQ1)}
We first confirm the accuracy phenomenon reported by prior subtype and
subpopulation-shift work, then turn to calibration.
Table~\ref{tab:gap-all-datasets} reports what each model does on unseen subtypes and
how far that falls from the seen split. Coarse accuracy drops by $29$--$40$ points on
CIFAR-100, IN-25 and BREEDS and by a milder $7$--$13$ on iNat-25, but confidence
drops by only $2$--$16$. On unseen subtypes accuracy lands at $30$--$70\%$ while
confidence stays at $64$--$90\%$. ECE therefore rises from $2$--$22$ points on seen
subtypes to $9$--$46$ on unseen, so miscalibration increases by $5$--$28$ points on
every model. The gap is not a binning artifact: it is unchanged to within $0.7$
points under equal-mass bins, and the binning-free NLL and Brier both rise from seen to
unseen on every row (Appendix~\ref{app:proper}).

\begin{figure*}[t]
\centering
\includegraphics[width=0.92\textwidth]{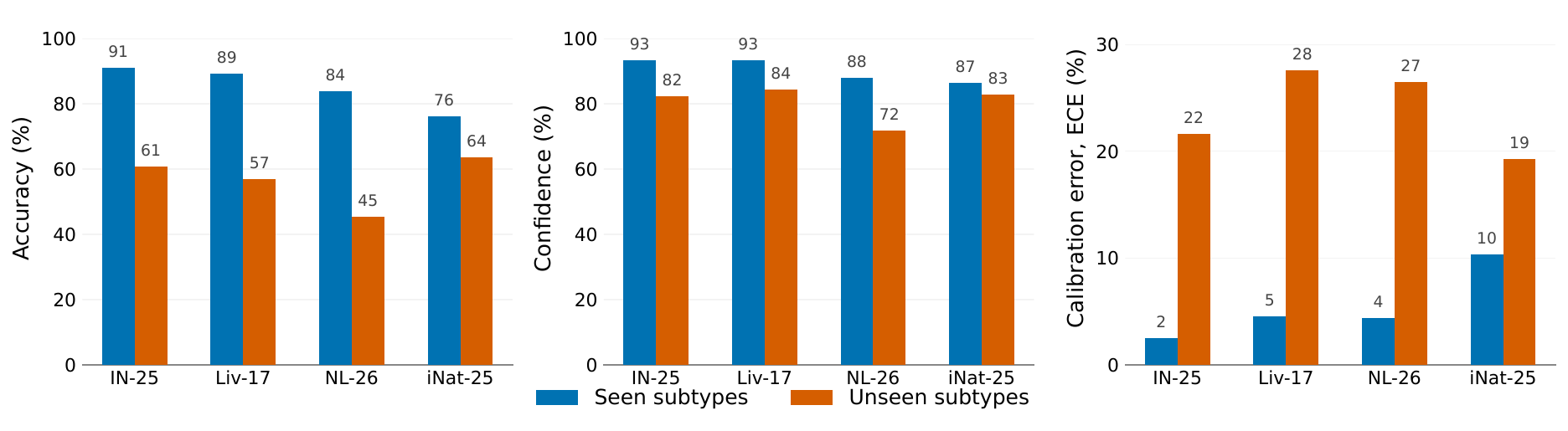}
\caption{Seen versus unseen subtypes (ResNet-50, direct-coarse from scratch). Accuracy drops far
more than confidence does, and the calibration error rises from $2$--$10$ points on seen subtypes
to $19$--$28$ on unseen.}
\label{fig:calibration-gap}
\end{figure*}

Figure~\ref{fig:calibration-gap} makes the dissociation visible for ResNet-50, one
metric per panel: accuracy falls steeply from seen to unseen subtypes, confidence
barely moves, and the calibration error rises on every dataset. Per-architecture reliability diagrams are in Appendix~\ref{app:reliability}.

Confidence also degrades as a per-sample signal. On unseen subtypes the
failure-detection AUROC falls on every dataset, by $0.15$--$0.18$ on CIFAR-100,
IN-25 and BREEDS and by $0.05$ on iNat-25, consistent with its much smaller
accuracy gap, and the risk--coverage AURC rises $2.3\times$ to $18\times$ (Appendix~\ref{app:discrim}).
Used to decide when to abstain, such a model lets through far more errors while
answering the same fraction of inputs. This part of the failure is harder to repair
than $\Delta_{\mathrm{cal}}$, because AUROC and AURC depend on the order of the
confidence scores rather than their values, and recalibration changes only the values.

\subsection{Confidence holds up under subtype shift, unlike under matched generic corruption (RQ2)}
\label{sec:generic-shift}
Miscalibration under distribution shift is already
known~\cite{ovadia2019can,hendrycks2018deep}; the question is whether subtype novelty
behaves like the shifts that literature studies. We compare against a matched
generic-shift arm built from the ImageNet-C
corruptions~\cite{hendrycks2019benchmarking}. We apply all $15$ of them, at all $5$
severities, to the seen-subtype images of every dataset, leaving the coarse labels
unchanged. Corrupting our own seen images rather than using the released ImageNet-C
copies matters here: the clean and corrupted arms are then literally the same images, so
the only thing that changes is image quality.

The comparison rests on two choices. First, accuracy loss must be matched: a corruption
that destroys more accuracy will naturally cost more confidence, so confidence drops are
only comparable at equal accuracy loss. We therefore interpolate each corruption along
its own severity curve to the accuracy loss of that dataset's subtype shift. Second, no
corruption is hand-picked: every corruption that can reach that accuracy loss contributes
one point, and we report the mean over all of them. Matching a single best-fitting
corruption instead would be fragile, since which one fits best changes with the severity
grid.

\begin{figure}[t]
\centering
\includegraphics[width=\linewidth]{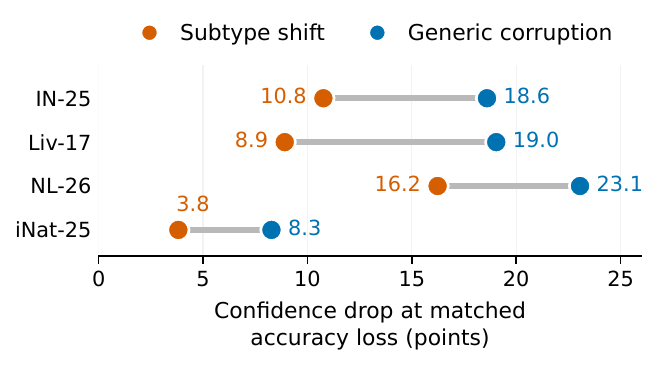}
\caption{Confidence given up at matched accuracy loss (ResNet-50, direct-coarse from scratch); the
corruption value is the mean over every ImageNet-C corruption that can reach that
dataset's subtype accuracy loss. Both arms use the clean accuracy of the same seen images
as reference, so the subtype gaps differ marginally from
Table~\ref{tab:gap-all-datasets}.}
\label{fig:rq2-corruption}
\end{figure}

A corruption enters only when its five severities bracket the subtype accuracy
loss, which leaves $14$ of the $15$ corruptions on IN-25, $11$ on each BREEDS
task, and $3$ on iNat-25, whose subtype gap is smaller than most corruptions cost even
at severity~1. Figure~\ref{fig:rq2-corruption} collects the surviving arms. At matched accuracy loss, generic
corruption costs $17.3$ points of confidence on average against $9.9$ for subtype shift
($18.6$ vs.\ $10.8$ on IN-25, $19.0$ vs.\ $8.9$ on Liv-17, $23.1$ vs.\ $16.2$ on
NL-26, $8.3$ vs.\ $3.8$ on iNat-25). The model loses the same accuracy either way, yet
on every dataset it gives up less confidence to the subtype shift: it reacts to visible
degradation and stays blind to in-taxonomy novelty. The corruption control is run on
ResNet-50; the dissociation between $\Delta_{\mathrm{acc}}$ and
$\Delta_{\mathrm{conf}}$ it builds on is present on all five architectures
(Table~\ref{tab:gap-all-datasets}).

\subsection{Post-hoc recalibration narrows but does not close the seen--unseen calibration gap (RQ3a)}
We tune the two post-hoc calibrators on a held-out image split from the seen subtypes and apply them to
unseen subtypes, a transfer that looks safe because the task and the coarse labels are
unchanged and only the set of fine subtypes differs.

\begin{figure}[b]
\centering
\includegraphics[width=\linewidth]{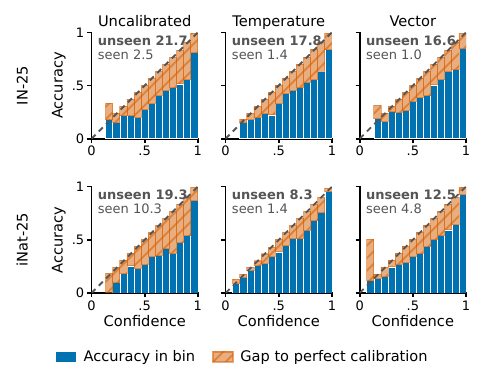}
\caption{Reliability diagrams on the unseen split (ResNet-50 direct-coarse from scratch,
$15$ equal-width bins; seen-fit temperature $T^*_{\mathcal S}=1.2$ on IN-25 and
$1.65$ on iNat-25). Bars below the diagonal are overconfident; each panel carries its
unseen and seen ECE.}
\label{fig:rq3-reliability}
\end{figure}

Table~\ref{tab:posthoc-calibration} gives the numbers and
Figure~\ref{fig:rq3-reliability} the shape, both for ResNet-50. Temperature scaling
lowers the unseen ECE on
every dataset, by $18$--$57\%$ (IN-25 $21.7\!\to\!17.8$, iNat-25
$19.3\!\to\!8.3$). Read on its own that looks like a repair, but what the calibrator
fixes is the split it was tuned on. On IN-25 the seen ECE falls to almost nothing
($2.5$ uncalibrated, $1.4$ under temperature, $1.0$ under vector scaling), while every
bin of the unseen split stays below the diagonal in
Figure~\ref{fig:rq3-reliability}. That leaves $\Delta_{\mathrm{cal}}$ at $7$--$17$. iNat-25 is
where recalibration helps most, and even there the calibrated unseen ECE ($8.3$) is six
times the calibrated seen ECE ($1.4$). The richer vector scaling improves the seen split
most and transfers worst to unseen subtypes (Liv-17 unseen ECE $25.3$ vs.\ $19.6$
under plain temperature),
because it may overfit the seen split it was tuned on. On every dataset the calibrated
unseen ECE stays far above the calibrated seen ECE. Recalibration has been reported to
transfer across hierarchy levels for vision-language
models~\cite{tu2024empiricalstudymatterscalibrating}. That transfer does not hold
under discriminative subtype shift. The same holds for all five architectures on all four datasets: temperature
scaling never drives $\Delta_{\mathrm{cal}}$ to zero (Appendix~\ref{app:arch}).

A temperature fit on the unseen split itself does close most of the gap, but it cannot be
chosen without knowing which inputs are unseen, which is what motivates the detection
question below (Appendix~\ref{app:oracle}).

\begin{table}[t]
\centering
\small
\setlength{\tabcolsep}{4pt}
\caption{Post-hoc calibration (ResNet-50, direct-coarse from scratch), in percentage points. Each calibrator is
tuned on a held-out image split from the seen subtypes and then applied to unseen subtypes;
\emph{None} is the uncalibrated model.}
\begin{tabular}{clccc}
\toprule
Dataset & Method &
$\mathrm{ECE}_{\mathcal{S}}\downarrow$ &
$\mathrm{ECE}_{\mathcal{U}}\downarrow$ &
$\Delta_{\mathrm{cal}}\downarrow$ \\
\midrule

\multirow[c]{3}{*}{IN-25}
  & None        & 2.5  & 21.7 & 19.2 \\
  & Temperature & 1.4  & 17.8 & 16.4 \\
  & Vector      & 1.0  & 16.6 & 15.6 \\

\midrule

\multirow[c]{3}{*}{Liv-17}
  & None        & 4.5  & 27.6 & 23.1 \\
  & Temperature & 2.2  & 19.6 & 17.4 \\
  & Vector      & 3.7  & 25.3 & 21.6 \\

\midrule

\multirow[c]{3}{*}{NL-26}
  & None        & 4.4  & 26.5 & 22.1 \\
  & Temperature & 1.6  & 18.7 & 17.1 \\
  & Vector      & 3.4  & 25.0 & 21.6 \\

\midrule

\multirow[c]{3}{*}{iNat-25}
  & None        & 10.3 & 19.3 & 9.0 \\
  & Temperature & 1.4  & 8.3  & 6.9 \\
  & Vector      & 4.8  & 12.5 & 7.7 \\

\bottomrule
\end{tabular}
\label{tab:posthoc-calibration}
\end{table}

\subsection{Inference-time scores flag novel subtypes only weakly (RQ3b)}
We ask whether a standard detection score can separate unseen-subtype from
seen-subtype inputs, so that a system could at least know when to distrust its own
confidence. We score every test input with the maximum softmax probability
(MSP)~\cite{hendrycks2017baseline}, the logit-based energy score~\cite{liu2020energy},
and the feature-based Mahalanobis distance to the nearest seen-class
centroid~\cite{lee2018simple,knowwhentoabstain2026,cedl2026}. Each score is evaluated on
its own by the \emph{novelty-detection} AUROC: the probability that a randomly drawn
unseen-subtype input is scored as more novel than a randomly drawn seen-subtype one, so
$0.5$ means the two populations are indistinguishable. This is a different question from
the failure-detection AUROC of RQ1, which asks whether confidence separates correct from
incorrect predictions inside a single split.

\begin{table}[b]
\centering
\small
\setlength{\tabcolsep}{10pt}
\caption{Novelty-detection AUROC of each score (ResNet-50, direct-coarse from scratch): how well it
ranks unseen-subtype above seen-subtype inputs, $0.5$~= indistinguishable. This is a
triage signal, not the failure-detection AUROC of Appendix~\ref{app:discrim} and not standard
out-of-taxonomy OOD rejection. Restricting each score to inputs classified correctly on
both splits lowers it by at most $0.04$ (Appendix~\ref{app:discrim}).}
\begin{tabular}{lccc}
\toprule
Dataset & MSP & Energy & Mahalanobis \\
\midrule
IN-25 &0.723 & 0.711 & 0.453 \\
Liv-17    & 0.723 & 0.698 & 0.456 \\
NL-26     & 0.740 & 0.738 & 0.463 \\
iNat-25   & 0.563 & 0.563 & 0.473 \\
\bottomrule
\end{tabular}
\label{tab:detection-auroc}
\end{table}

All three scores separate the two populations poorly
(Table~\ref{tab:detection-auroc}). The two output-based scores, MSP and energy, reach
only ${\sim}0.70$ on the three larger benchmarks and $0.56$ on iNat-25. That is above
chance but far from usable: at a threshold that flags $5\%$ of seen-subtype inputs, MSP
catches only $12$--$16\%$ of unseen-subtype inputs ($6\%$ on iNat-25), because the two
score distributions overlap heavily. The feature-based Mahalanobis distance sits just
below chance on every dataset ($0.45$--$0.47$), ranking unseen subtypes as slightly
\emph{less} novel than seen ones. A network trained only on the coarse label collapses
same-supertype subtypes together, so an unseen subtype ends up looking like the seen
ones in feature space.

Confidence correlates with correctness, so this separation could be trivial error
detection rather than novelty detection. Restricting each score to inputs the model
classifies correctly on both splits lowers MSP and energy by at most $0.04$
(Appendix~\ref{app:discrim}), so the signal does respond to novelty, but far too weakly to act on. No
architecture does better (Appendix~\ref{app:arch}). On a
from-scratch model, then, confidence does not track subtype novelty, post-hoc
recalibration does not remove it, and inference-time detection surfaces it only weakly.

\subsection{Calibration collapse varies by superclass and tracks seen-subtype diversity}
The per-class collapse is highly heterogeneous: some coarse classes keep
near-perfect calibration on unseen subtypes, others collapse completely (per-supertype
accuracy and ECE in Appendix~\ref{app:percat}). This matches the mechanism of
\citet{lyu2025imbalance}: under-represented groups receive inflated training
logits, and so become overconfident without losing accuracy. Subtype shift is the
extreme case, since unseen subtypes are absent from training rather than merely
rare.

\begin{figure}[t]
\centering
\includegraphics[width=\linewidth]{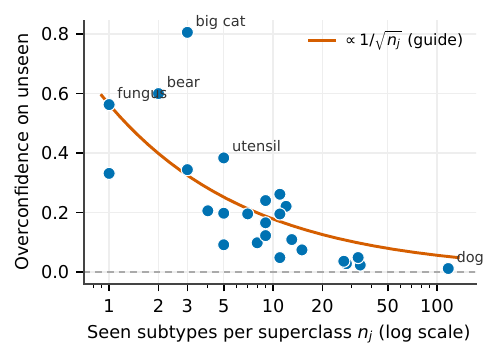}
\caption{Per-supertype overconfidence (mean confidence $-$ accuracy on unseen subtypes)
against the number of seen subtypes $n_j$ (IN-25, ResNet-50 direct-coarse from
scratch). Each point is one of
the $25$ supertypes; the curve is a $1/\sqrt{n_j}$ guide (Spearman $\rho=-0.82$).}
\label{fig:imagenet-subtype}
\end{figure}

We ask whether the collapse is predictable from seen-side signals. The one that tracks it is the number of seen subtypes per coarse class, $n_j=|\mathcal{S}_j|$. On IN-25 (the only dataset where it varies), supertypes with few seen subtypes are the most overconfident on unseen ones while richly-sampled ones (e.g.\ \textit{dog}, 117 subtypes) stay calibrated (Figure~\ref{fig:imagenet-subtype}; Spearman $\rho=-0.82$), consistent with a sublinear ($\approx\!1/\sqrt{n_j}$) coverage rate~\cite{hierambsets2026}. The trend is not a difficulty artifact: regressing per-supertype overconfidence on $1/\sqrt{n_j}$ and seen accuracy shows $1/\sqrt{n_j}$ to be the dominant predictor ($t=4.2$), while seen accuracy carries no signal ($t=0.0$).

A controlled $k$-sweep supports a causal reading: holding the per-class image budget fixed and varying only the seen-subtype count shrinks $\Delta_{\mathrm{cal}}$ monotonically from $31$ ($k{=}1$) to $23$ ($k{=}4$; Appendix~\ref{app:ksweep}).

The count is useful in aggregate but not per class. Ranking supertypes by it and keeping only the better-sampled half roughly halves the overconfidence in the retained supertypes ($10.2$, against $21.6$ across all supertypes), so it works as a coarse triage signal. It does not say which individual supertype will collapse, and the two obvious alternatives, feature dispersion and confidence margin, carry no signal at all.

Initialization matters more than protocol. Switching to the two-stage fine-pretrained-frozen protocol keeps $\Delta_{\mathrm{cal}}$ within $6$ points, while an ImageNet-1K-pretrained backbone lowers it from $19.2$ to $12.8$ on IN-25 without closing it (Appendix~\ref{app:protocol}).

\subsection{Discussion and Future Work}
The practical upshot is that subtype robustness must be audited with calibration, since
accuracy alone hides the failure. A model that loses accuracy on novel subtypes goes on
reporting almost the same confidence there, and neither post-hoc recalibration nor
inference-time detection recovers the lost calibration. The two failure modes are complementary: an
aggregate miscalibration that a single temperature could repair only with oracle
knowledge of which inputs are unseen, and a per-sample discrimination loss that no
monotone rescaling can touch.
Post-hoc recalibration addresses the first and inference-time detection the second, so
neither reaches this regime.

Subtype shift also behaves differently from generic covariate shift. At matched accuracy
loss, visible corruption reduces confidence much more sharply, whereas confidence barely
moves on unseen subtypes even though the inputs look ordinary and still have a valid
coarse label (RQ2). Together with the loss of per-sample discrimination on unseen
subtypes (RQ1), this shows that subtype robustness has been evaluated too narrowly.
Preserving coarse-label accuracy is not enough if confidence no longer reflects
reliability, so subtype robustness should be treated as a joint requirement on prediction
and confidence.

We study image classification, and we isolate the phenomenon rather than propose a
training method, so several directions stay open. The most direct is dense prediction,
where subtype shift acts at the pixel or region level in semantic segmentation and object
detection. Beyond vision, whether the same silent overconfidence arises in language or
tabular models is untested, and so is whether foundation and vision-language models
behave differently. Our one pretraining control points only to partial mitigation
(Appendix~\ref{app:protocol}), where an ImageNet-pretrained backbone narrows the gap without closing it.

Three method-side handles are worth pursuing. Training-time objectives such as evidential
learning, outlier exposure, or subtype-diversity augmentation would attack the
miscalibration at its source. A detector designed for in-taxonomy novelty rather than
out-of-taxonomy rejection would give the triage signal that the off-the-shelf scores of
RQ3b fail to provide. A larger diversity sweep than our $k{=}1$ to $4$ would turn the
seen-subtype effect from a supported direction into a quantitative law. Until such tools
exist, the immediate evaluation requirement is clear: confidence reliability should be
reported alongside coarse accuracy whenever subtype robustness is claimed.

\paragraph{Generative AI use statement.}
Generative AI tools were used solely for language editing and suggestions on
the organization of the exposition. All research questions, experimental
design, analyses, results, and conclusions were developed by the authors.
The authors reviewed and verified all AI-assisted edits, including all
references, and take full responsibility for the content of the manuscript.

\section*{Acknowledgments}
This work used Jetstream2 at Indiana University through ACCESS allocation
CIS260254 from the Advanced Cyberinfrastructure Coordination Ecosystem:
Services \& Support (ACCESS) program, which is supported by U.S. National
Science Foundation grants \#2138259, \#2138286, \#2138307, \#2137603, and
\#2138296. Results were also obtained using the Chameleon testbed, supported by
the National Science Foundation. We thank the Jetstream2, ACCESS, and Chameleon
support teams for the computational infrastructure used in this work. We also
thank Yutong Wang for helpful discussions during the early stages of this
project.

%======================= REFERENCES =======================
% \newpage, not \pagebreak: the style sets \flushbottom, and \pagebreak lets TeX
% stretch the inter-paragraph glue to fill this short last body page.
\newpage
\label{bodyend}%
\bibliography{references}

%======================= APPENDIX =======================
\clearpage
\appendix
\section{Appendix}

\appsec{app:datasets}{Datasets: final sizes, splits, and construction.}
All five benchmarks share one recipe: a coarse \emph{supertype} valid on both train and test, with its fine
\emph{subtypes} partitioned into \emph{seen} (training) and \emph{unseen} (test-only); the model predicts the
supertype. We use a strict \textbf{four-way split}: \emph{train} / \emph{seen-val} / \emph{test-seen} /
\emph{test-unseen}. \emph{train}, \emph{seen-val} and \emph{test-seen} are disjoint sets of \emph{images} carved
from the same \emph{seen} subtypes, so no subtype label is withheld from training; \emph{seen-val} is used both
for model selection and for fitting the post-hoc calibrators of RQ3a (never fit on test).
\emph{test-unseen} holds the images of the \emph{unseen} subtypes. \textbf{No images are copied}: each split is
a directory of symbolic links into a single source copy of the dataset, and the exact partition is frozen in a
seeded \texttt{split\_manifest.json}, the open-sourceable artifact: we release the subtype lists and the
deterministic construction code (reproducible from \texttt{-{}-seed} alone), not the images. Models (AlexNet,
ResNet-18, ResNet-50, ViT-B/16 and ConvNeXt-T; CIFAR-100 uses only AlexNet and ResNet-18) are trained
from scratch on the supertype label; all metrics are computed on a per-supertype-balanced test set with
equal-width ECE (Appendix~\ref{app:protocol} reports a second-seed check). Table~\ref{tab:datasets-final} gives the final sizes
and Table~\ref{tab:training-recipe} the full training and evaluation recipe.

\begin{table*}[tbp]\centering\small
\caption{The five subtype-shift benchmarks: construction, splits, and protocol. One schema throughout:
predict the coarse \emph{supertype}, train only on \emph{seen} subtypes, test on both seen and \emph{unseen}
subtypes of the same supertype. ``subt.\ (s/u)''~= fine subtypes seen/unseen; image counts are the
built subset sizes. ``Who splits; protocol'' gives the partition source and how the supertype classifier
is built: \emph{direct-coarse} (end-to-end on supertype labels, from scratch) or
\emph{fine-pretrained frozen} (train on seen subtypes, freeze, learn a supertype head). The IN-25 split is
deliberately imbalanced across supertypes (e.g.\ \textit{dog} 117 vs \textit{fungus} 1 seen subtype), inherited from
ImageNet-1K; we do not rebalance it and instead evaluate on a per-supertype-balanced test set. All five
benchmarks are built and the full train$\to$logits$\to$calibration pipeline is validated end-to-end. Train
counts are the seen set after the val holdout; IN-25's train pool is $\subset$488{,}644 and all test
sizes are the per-supertype-balanced evaluation sets.}
\label{tab:datasets-final}
\begin{tabular}{lccrrrl}
\toprule
Dataset & \#super & subt.\ (s/u) & train & test-seen & test-unseen & Who splits; protocol \\
\midrule
CIFAR-100   & 20 & 60/40           & 30{,}000        & 6{,}000  & 4{,}000  & us ($k{=}3/2$); direct-coarse (scratch) \\
IN-25 & 25 & 383/3{,}872     & $\sim$488{,}644 & 7{,}179  & 7{,}500  & Hendrycks; direct-coarse (scratch) \\
Liv-17      & 17 & 34/34           & 39{,}780        & 1{,}700  & 1{,}700  & BREEDS; direct-coarse (scratch) \\
NL-26       & 26 & 52/52           & 60{,}108        & 2{,}600  & 2{,}600  & BREEDS; direct-coarse (scratch) \\
iNat-25     & 25 & 3{,}427/3{,}427 & 171{,}000       & 34{,}000 & 34{,}000 & us (order); direct-coarse (scratch) \\
\bottomrule
\end{tabular}
\end{table*}

\begin{table*}[tbp]
\centering
\small
\caption{Training and evaluation recipe. ResNet-50 is the primary backbone; all
five architectures are swept on IN-25 and BREEDS, while CIFAR-100 uses AlexNet
and ResNet-18. ``Init'' and ``Protocol'' are the two training axes. Hyper-parameters marked ``seen-VAL'' are tuned
per dataset on the held-out seen-subtype validation split, not fixed a priori.
The calibration metrics (proper scores, equal-width ECE, per-supertype balancing)
are defined in the Evaluation section. Post-hoc calibrators
and detection scores are fit and applied only at evaluation time on saved logits.}
\begin{tabular}{l p{0.74\textwidth}}
\toprule
Component & Setting \\
\midrule
Backbones (sweep)      & AlexNet, ResNet-18, \textbf{ResNet-50} (primary), ViT-B/16, ConvNeXt-T; CIFAR-100 uses only AlexNet and ResNet-18 \\
Init                   & from scratch (all main-table datasets); an ImageNet-1K-pretrained backbone on IN-25 is a separate initialization control \\
Protocol               & direct-coarse (all main-table datasets); the from-scratch fine-pretrained-frozen variant (CIFAR-100, BREEDS) is a separate control (Appendix~\ref{app:protocol}) \\
Input resolution       & 32 (CIFAR-100, except AlexNet/ViT-B/16 upsampled to 224); 224 (BREEDS, iNat-25, IN-25) \\
Optimizer              & SGD (momentum 0.9) for ResNet/AlexNet; AdamW ($\beta{=}(0.9,0.999)$, linear warmup) for ViT-B/16 and ConvNeXt-T, which diverge under SGD; cosine LR decay; weight decay $10^{-4}$--$5\times10^{-4}$ (SGD) or $0.05$ (AdamW) \\
Learning rate / batch  & lr $0.1$ (SGD); $10^{-3}$/$3\times10^{-4}$ (AdamW ViT/ConvNeXt); batch 128, reduced to 96 on the memory-limited IN-25 transformer runs \\
Model selection        & per-dataset epoch budget (BREEDS, iNat-25 90; IN-25 50; CIFAR-100 120--150 backbone $+$ 40--50 head); cosine schedule; \emph{best-on-seen-VAL} checkpoint kept, not the last epoch \\
Seeds                  & the seen/unseen subtype split and the recipe are fixed, so a seed re-initializes the model only; every reported number is a single run at seed $42$, and a second-seed robustness check is reported in Appendix~\ref{app:protocol} \\
Compute                & one NVIDIA A100 (GRID A100X-20C vGPU, $20$\,GB), $16$ CPU cores, $58$\,GB RAM; Ubuntu 24.04 LTS; Python 3.12.3, PyTorch 2.12.1 (CUDA 12.6), torchvision 0.27.1, cuDNN 9.10.2, NumPy 2.4.4 \\
\midrule
Accuracy               & coarse (supertype) accuracy, seen vs.\ unseen \\
Calibration            & NLL, Brier (proper scores, primary); equal-width ECE, per-supertype balanced \\
Discrimination         & failure-detection AUROC and risk-coverage AURC (seen vs.\ unseen) \\
Post-hoc calibrators   & temperature scaling, vector scaling (fit on seen-VAL, applied to unseen); oracle-$T$ residual \\
Detection scores       & MSP, energy, Mahalanobis (overall and within-correct) \\
\bottomrule
\end{tabular}
\label{tab:training-recipe}
\end{table*}

\appsec{app:protocol}{Protocol and initialization controls.}
The main table fixes two axes (direct-coarse, from scratch). We vary each as a
control fit, like the main results, on a held-out seen-val.
\textbf{Protocol is neutral} (Table~\ref{tab:app-protocol}): across CIFAR-100 and both
BREEDS tasks, switching one-stage direct-coarse for two-stage
fine-pretrained-frozen leaves the unseen calibration gap essentially unchanged
($|\Delta\Delta_{\mathrm{cal}}|<6$, small against gaps of $20$--$30$) at matched
accuracy, so the two-stage protocol neither creates nor repairs silent overconfidence. \emph{Pretraining reduces but does not remove the
gap} (Table~\ref{tab:app-init}): on IN-25 an ImageNet-1K-pretrained frozen
backbone lowers $\Delta_{\mathrm{cal}}$ ($19.2\!\to\!12.8$) without closing it, and it
does so by making confidence itself more responsive (at essentially unchanged
$\Delta_{\mathrm{acc}}$, $\Delta_{\mathrm{conf}}$ rises from $11.2$ to $14.5$), while
it brings the subtype and generic-corruption arms together (their
$\Delta_{\mathrm{conf}}/\Delta_{\mathrm{acc}}$ ratios become close) and makes unseen
subtypes detectable (Mahalanobis $0.45\!\to\!0.75$), because its 1K features displace
the 22K-only unseen synsets in feature space. These controls unify: the from-scratch
coarse features are blind to in-taxonomy novelty (the dissociation holds and detection
fails), whereas an ImageNet-pretrained backbone partly sees it: confidence now
falls somewhat under subtype shift (narrowing the dissociation) and the unseen synsets
become separable. Yet the overconfidence is only reduced, not removed. Pretraining is
thus a partial mitigation, and the protocol axis changes nothing.

\paragraph{Seed sensitivity.}
The seen/unseen subtype split, the data, and the per-architecture recipe are fixed, so a
model-init seed changes only the weight initialization. As a robustness check we
retrained a second seed and found the calibration gaps unchanged in sign and closely
matched in magnitude.

\begin{table*}[tbp]
\centering\small
\setlength{\tabcolsep}{5pt}
\caption{Protocol control: one-stage direct-coarse vs.\ two-stage fine-pretrained-frozen,
at matched accuracy, across three datasets (five architectures on each BREEDS task, the
two small ones on CIFAR-100). All gaps are in percentage points. Confidence stays as unresponsive under one protocol as the other, and
the unseen calibration gap barely moves ($|\Delta\Delta_{\mathrm{cal}}|<6$, against gaps
of $20$--$30$), so the protocol neither creates nor repairs silent overconfidence.}
\label{tab:app-protocol}
\begin{tabular}{ll cc cc cc c}
\toprule
 & & \multicolumn{2}{c}{$\Delta_{\mathrm{acc}}$} & \multicolumn{2}{c}{$\Delta_{\mathrm{conf}}$} & \multicolumn{2}{c}{$\Delta_{\mathrm{cal}}$} & \\
\cmidrule(lr){3-4}\cmidrule(lr){5-6}\cmidrule(lr){7-8}
Dataset & Model & 1-stage & 2-stage & 1-stage & 2-stage & 1-stage & 2-stage & $\Delta\Delta_{\mathrm{cal}}$ \\
\midrule
\multirow[c]{2}{*}{CIFAR-100}
  & AlexNet    & 34.8 & 33.6 & 7.2 & 6.8 & 27.6 & 26.8 & $-0.8$ \\
  & ResNet-18  & 39.9 & 39.9 & 12.4 & 11.6 & 27.5 & 28.3 & $+0.8$ \\
\midrule
\multirow[c]{5}{*}{Liv-17}
  & AlexNet    & 33.0 & 33.8 & 9.8 & 9.4 & 22.5 & 24.4 & $+1.9$ \\
  & ResNet-18  & 28.7 & 29.7 & 8.0 & 8.8 & 21.0 & 20.3 & $-0.6$ \\
  & ResNet-50  & 32.4 & 31.6 & 8.9 & 9.0 & 23.1 & 22.7 & $-0.4$ \\
  & ViT-B/16   & 37.2 & 34.9 & 9.2 & 9.6 & 28.0 & 25.2 & $-2.8$ \\
  & ConvNeXt-T & 32.1 & 32.9 & 7.3 & 9.0 & 24.7 & 23.9 & $-0.7$ \\
\midrule
\multirow[c]{5}{*}{NL-26}
  & AlexNet    & 35.9 & 38.6 & 15.8 & 15.4 & 20.0 & 23.0 & $+3.0$ \\
  & ResNet-18  & 37.6 & 38.3 & 15.4 & 16.1 & 22.0 & 22.0 & $-0.0$ \\
  & ResNet-50  & 38.6 & 38.2 & 16.2 & 15.9 & 22.1 & 22.0 & $-0.2$ \\
  & ViT-B/16   & 32.9 & 30.2 & 8.5 & 11.1 & 24.5 & 19.1 & $-5.4$ \\
  & ConvNeXt-T & 38.0 & 36.2 & 13.2 & 14.7 & 24.8 & 21.4 & $-3.3$ \\
\bottomrule
\end{tabular}
\end{table*}

\begin{table}[htbp]
\centering\small
\setlength{\tabcolsep}{6pt}
\caption{Initialization control (IN-25, ResNet-50): from-scratch direct-coarse
vs.\ ImageNet-1K-pretrained frozen. Gaps are in percentage points. Pretraining reduces
but does not remove the unseen calibration gap ($\Delta_{\mathrm{cal}}$), and it does so
by making confidence more responsive: at essentially unchanged $\Delta_{\mathrm{acc}}$,
$\Delta_{\mathrm{conf}}$ rises from $11.2$ to $14.5$. It also brings the subtype and
generic-corruption arms together (their $\Delta_{\mathrm{conf}}/\Delta_{\mathrm{acc}}$
ratios become close) and makes unseen subtypes detectable (Mahalanobis
novelty-detection AUROC).}
\label{tab:app-init}
\begin{tabular}{lcc}
\toprule
 & From scratch & Pretrained \\
 & (direct-coarse) & (frozen) \\
\midrule
Accuracy gap $\Delta_{\mathrm{acc}}$    & 30.3   & 28.0 \\
Confidence gap $\Delta_{\mathrm{conf}}$ & 11.2   & 14.5 \\
Calibration gap $\Delta_{\mathrm{cal}}$ & $+19.2$ & $+12.8$ \\
Subtype ratio                           & 0.37   & 0.52 \\
Generic ratio                           & 0.76   & 0.56 \\
Mahalanobis novelty AUROC               & 0.453  & 0.747 \\
\bottomrule
\end{tabular}
\end{table}

\appsec{app:binning}{The calibration gap is robust to the binning scheme.}
The standard equal-width estimator~\cite{naeini2015obtaining} partitions \([0,1]\) into
bins \(\{I_b\}_{b=1}^B\) with \(B_b=\{i:c(x_i)\in I_b\}\); on a split \(T\),
\begin{align}
\widehat{\mathrm{ECE}}(T)
&=\sum_{b=1}^{B}\frac{|B_b|}{|T|}
\left|\widehat{\mathrm{acc}}(B_b)-\widehat{\mathrm{conf}}(B_b)\right|,
\label{eq:ece-empirical}
\end{align}
where empty bins contribute zero. This estimator is biased, and the bias grows as
samples-per-bin shrink, so we never read a calibration gap off ECE alone. Our
confidences are also heavily concentrated: on IN-25, $80\%$ of seen-split and
$50\%$ of unseen-split predictions fall in the single top equal-width bin, exactly the
regime in which equal-width ECE is criticised as unstable. Appendix~\ref{app:proper} therefore
recomputes every gap under equal-mass bins and against two binning-free proper scores.

\appsec{app:proper}{The calibration gap does not depend on the ECE histogram.}
Table~\ref{tab:gap-full} recomputes every gap in the main table
(Table~\ref{tab:gap-all-datasets}) two ways. First under equal-mass (adaptive) bins,
which place roughly equal counts per bin and so resolve the concentrated
high-confidence region finely: the two binning schemes agree to within $0.7$ points on
every row. Second with two binning-free proper scores~\cite{gneiting2007proper}, the
negative log-likelihood
\(\mathrm{NLL}=-\tfrac{1}{|T|}\sum_i \log p_{\supertype{y_i}}(x_i)\) and the Brier score
\(\tfrac{1}{|T|}\sum_i\sum_{j}(p_j(x_i)-\mathbf{1}\{\supertype{y_i}=j\})^2\); both
rise from seen to unseen on all $22$ rows, and their rise tracks
$\Delta_{\mathrm{cal}}$ closely (Spearman $\rho=0.96$ and $0.95$). The gap is therefore
a property of the predicted probabilities, not of where the bin edges fall.

\begin{table}[htbp]
\centering
\small
\setlength{\tabcolsep}{2pt}
\caption{The calibration gap under a second binning scheme and under two binning-free
proper scores. $\Delta_{\mathrm{cal}}$ is in percentage points; $\Delta$NLL and
$\Delta$Brier are the unseen minus seen rise in each score. All three move together on
every row.}
\label{tab:gap-full}
\begin{tabular}{ll cc cc}
\toprule
 & & \multicolumn{2}{c}{$\Delta_{\mathrm{cal}}$} & \multicolumn{2}{c}{Proper scores} \\
\cmidrule(lr){3-4}\cmidrule(lr){5-6}
Dataset & Model & eq.-width & eq.-mass & $\Delta$NLL & $\Delta$Brier \\
\midrule
\multirow[c]{2}{*}{CIFAR-100}
  & AlexNet    & 27.6 & 27.5 & 2.99 & 0.580 \\
  & ResNet-18  & 27.5 & 27.5 & 2.14 & 0.630 \\
\midrule
\multirow[c]{5}{*}{IN-25}
  & AlexNet    & 19.3 & 19.3 & 1.34 & 0.425 \\
  & ResNet-18  & 18.6 & 18.7 & 1.44 & 0.458 \\
  & ResNet-50  & 19.2 & 19.2 & 1.46 & 0.463 \\
  & ViT-B/16   & 25.4 & 25.4 & 2.18 & 0.561 \\
  & ConvNeXt-T & 21.3 & 21.3 & 1.72 & 0.491 \\
\midrule
\multirow[c]{5}{*}{Liv-17}
  & AlexNet    & 22.5 & 23.2 & 1.87 & 0.527 \\
  & ResNet-18  & 21.0 & 20.7 & 1.74 & 0.478 \\
  & ResNet-50  & 23.1 & 23.4 & 1.84 & 0.519 \\
  & ViT-B/16   & 28.0 & 28.1 & 2.42 & 0.617 \\
  & ConvNeXt-T & 24.7 & 24.8 & 2.03 & 0.548 \\
\midrule
\multirow[c]{5}{*}{NL-26}
  & AlexNet    & 20.0 & 20.0 & 1.68 & 0.498 \\
  & ResNet-18  & 22.0 & 22.2 & 2.01 & 0.558 \\
  & ResNet-50  & 22.1 & 22.4 & 1.95 & 0.564 \\
  & ViT-B/16   & 24.5 & 24.5 & 2.38 & 0.522 \\
  & ConvNeXt-T & 24.8 & 24.8 & 2.21 & 0.572 \\
\midrule
\multirow[c]{5}{*}{iNat-25}
  & AlexNet    & 5.1 & 5.1 & 0.32 & 0.098 \\
  & ResNet-18  & 8.5 & 8.5 & 0.70 & 0.198 \\
  & ResNet-50  & 9.0 & 9.0 & 0.72 & 0.199 \\
  & ViT-B/16   & 7.2 & 7.2 & 0.74 & 0.154 \\
  & ConvNeXt-T & 7.7 & 7.7 & 0.73 & 0.166 \\
\bottomrule
\end{tabular}
\end{table}

\appsec{app:construction}{Per-dataset construction details.}
We detail here the construction of \emph{iNaturalist} (iNat-25)~\cite{vanhorn2021inaturalist},
whose real biological taxonomy broadens coverage along an axis the other main datasets cannot
reach. It is evaluated with the identical pipeline used in the main text: train a
coarse (supertype) classifier on seen subtypes, then measure accuracy and calibration on
held-out (unseen) subtypes, so ECE/NLL/Brier/$\Delta_{\mathrm{cal}}$ are directly
comparable.

\textbf{iNaturalist.} From iNat2021 ($10{,}000$ species, $2.7$M images; we use the official
$50$-image/species \texttt{train\_mini} subset for training and the $10$/species \texttt{val}
split for evaluation) we take a biological rank as the supertype (\textsc{order}: the $25$
orders with $\geq\!100$ species, spanning three kingdoms; or \textsc{family} for finer,
better-balanced supertypes) and split the species under each supertype into seen vs.\
unseen. The $25$-order subset covers $\approx\!6{,}854$ species, giving $\approx\!171$K
training images ($3{,}427$ seen species $\times 50$) and $\approx\!34$K images per test split
($\times 10$); order imbalance is handled by macro-averaging rather than capping. A ResNet-50 is trained from scratch to predict
the supertype on seen-species images; held-out species (same supertypes) form the test
split. Since the supertype is a true taxonomic node, the seen$\to$unseen \emph{taxonomic
distance} (same genus / family / order) is a controlled severity knob. Ready-made species
partitions are available from prior work~\cite{wallin2025prohoc,wallin2026semihoc}.

\textbf{Protocol.} ResNet-50 from scratch to avoid pretraining
leakage; per-supertype-balanced accuracy and ECE (adaptive binning); the two post-hoc
calibrators, seen-side risk predictors, and confidence-based detection computed
exactly as in the main text. A matched generic-corruption
(covariate-shift) arm is run to confirm that the calibration failure follows semantic
subtype novelty rather than appearance degradation.

\begin{figure*}[tbp]
\centering
\includegraphics[width=\linewidth]{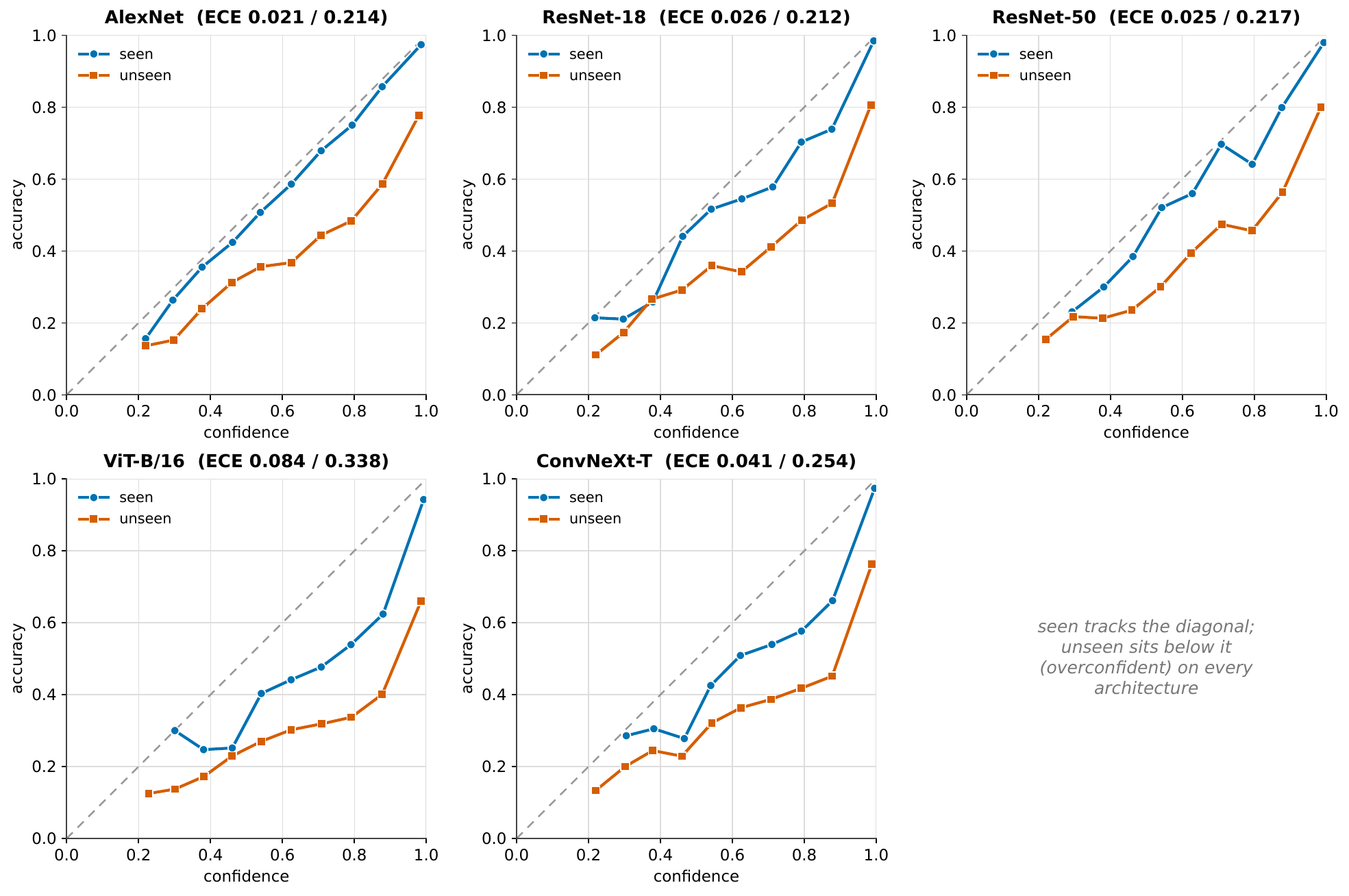}
\caption{Reliability diagrams (IN-25, from scratch, direct-coarse; ResNet-50 and
four other backbones): the unseen curve falls below the diagonal (overconfident) on every
architecture, and temperature scaling would slide each curve along, not onto, the
diagonal.}
\label{fig:app-reliability}
\end{figure*}

\appsec{app:reliability}{Full reliability diagrams.}
Figure~\ref{fig:app-reliability} draws reliability curves (binned confidence vs.\
empirical accuracy) for all five architectures on IN-25, seen and unseen. The seen
curves track the diagonal; the unseen curves sit well below it (overconfident) on every
architecture, so the gap is not an artifact of any one backbone.

\appsec{app:discrim}{Per-sample discrimination: failure-detection AUROC and risk--coverage AURC.}
Calibration is an aggregate property and does not say whether confidence still ranks
individual predictions usefully, so we report two per-sample measures on each
split (Table~\ref{tab:app-discrim}). The \emph{failure-detection AUROC} asks whether
$c(x)$ separates correct from incorrect predictions within a split: it is the
probability that a randomly drawn correct prediction is more confident than a randomly
drawn incorrect one ($1$~= perfect, $0.5$~= chance). The AURC is the area under
the risk--coverage curve obtained by answering only the most confident fraction of
inputs and abstaining on the rest (lower is better), so a rise means the same
confidence-thresholded policy now incurs more error at equal coverage.

Both degrade on unseen subtypes for every dataset, and the magnitude follows the
severity of the shift: the drop is $+0.15$--$0.18$ on CIFAR-100, IN-25 and BREEDS,
whose accuracy gaps are $29$--$40$ points, but only $+0.05$ on iNat-25, whose
accuracy gap is $12.6$; AURC rises $2.3\times$ (iNat-25) to $18\times$ (IN-25).
AUROC is rank-based and therefore invariant to any monotone rescaling of confidence:
unlike the calibration gap, this degradation cannot be repaired by temperature scaling.

\paragraph{Novelty or error detection?}
The novelty-detection AUROCs of Table~\ref{tab:detection-auroc} could in principle be
trivial error detection: accuracy is lower on unseen subtypes, and confidence
correlates with correctness, so a score that merely ranks errors would separate the two
splits. Restricting each score to inputs the model classifies correctly on both
splits removes that confound. The separation survives, falling by at most $0.04$: MSP
$0.723\!\to\!0.684$ (IN-25), $0.723\!\to\!0.701$ (Liv-17), $0.740\!\to\!0.714$
(NL-26) and $0.563\!\to\!0.538$ (iNat-25), with energy behaving the same way
($0.711\!\to\!0.681$, $0.698\!\to\!0.680$, $0.738\!\to\!0.702$,
$0.563\!\to\!0.545$). Mahalanobis is already at
chance, so its within-correct value is within noise and omitted. The weak signal is
therefore a genuine response to subtype novelty rather than an artifact of correctness.

\begin{table}[htbp]
\centering\small
\setlength{\tabcolsep}{3pt}
\caption{Per-sample discrimination (ResNet-50; ResNet-18 for CIFAR-100, matching the
main table), seen vs.\ unseen. Failure-detection AUROC (higher is better, $0.5$~=
chance) and risk--coverage AURC (lower is better).}
\label{tab:app-discrim}
\setlength{\tabcolsep}{2.4pt}
\begin{tabular}{lcccccc}
\toprule
 & & \multicolumn{3}{c}{failure-detection AUROC} & \multicolumn{2}{c}{AURC} \\
\cmidrule(lr){3-5}\cmidrule(lr){6-7}
Dataset & $\Delta_{\mathrm{acc}}$ & seen $\uparrow$ & unseen $\uparrow$ & $\Delta$ $\downarrow$ & seen $\downarrow$ & unseen $\downarrow$ \\
\midrule
CIFAR-100 & 39.9 & 0.894 & 0.733 & +0.161 & 0.031 & 0.363 \\
IN-25 &30.3 & 0.931 & 0.783 & +0.148 & 0.011 & 0.196 \\
Liv-17    & 32.4 & 0.928 & 0.748 & +0.180 & 0.014 & 0.248 \\
NL-26     & 38.6 & 0.917 & 0.757 & +0.160 & 0.028 & 0.331 \\
iNat-25   & 12.6 & 0.895 & 0.846 & +0.049 & 0.058 & 0.136 \\
\bottomrule
\end{tabular}
\end{table}

\appsec{app:arch}{Do the RQ3 conclusions depend on the architecture?}
The main text reports RQ3a and RQ3b on ResNet-50. Since both only need saved logits,
we can repeat them for all five architectures at no additional training cost
(Table~\ref{tab:app-arch-rq3}). The conclusions do not move. Seen-tuned temperature
scaling reduces \(\Delta_{\mathrm{cal}}\) on every one of the $20$ (dataset,
architecture) pairs but never to zero, leaving $3.3$ to $19.4$ points; vector scaling
leaves $5.1$ to $23.4$. Novelty detection stays weak everywhere, with the maximum
softmax probability between $0.52$ and $0.75$ AUROC. The one pattern the ResNet-50 row
hides is that ViT-B/16 has the \emph{largest} gap on all three ImageNet-derived
benchmarks, both before and after recalibration, so a more modern backbone is not the
safer choice here.

\begin{table}[htbp]
\centering\small
\setlength{\tabcolsep}{1.4pt}
\caption{RQ3 across architectures (direct-coarse from scratch). $\Delta_{\mathrm{cal}}$
in percentage points under no calibration, temperature scaling and vector scaling;
novelty-detection AUROC for the two output-based scores.}
\label{tab:app-arch-rq3}
\begin{tabular}{ll ccc cc}
\toprule
 & & \multicolumn{3}{c}{$\Delta_{\mathrm{cal}}\downarrow$} & \multicolumn{2}{c}{Novelty AUROC} \\
\cmidrule(lr){3-5}\cmidrule(lr){6-7}
Dataset & Model & None & Temp. & Vec. & MSP & Energy \\
\midrule
\multirow[c]{5}{*}{IN-25}
  & AlexNet    & 19.3 & 17.6 & 13.2 & 0.644 & 0.631 \\
  & ResNet-18  & 18.6 & 15.5 & 15.1 & 0.717 & 0.706 \\
  & ResNet-50  & 19.2 & 16.4 & 15.6 & 0.723 & 0.711 \\
  & ViT-B/16   & 25.4 & 19.3 & 19.8 & 0.691 & 0.695 \\
  & ConvNeXt-T & 21.3 & 16.9 & 17.5 & 0.727 & 0.723 \\
\midrule
\multirow[c]{5}{*}{Liv-17}
  & AlexNet    & 22.5 & 18.6 & 20.8 & 0.715 & 0.695 \\
  & ResNet-18  & 21.0 & 16.3 & 19.7 & 0.728 & 0.697 \\
  & ResNet-50  & 23.1 & 17.4 & 21.6 & 0.723 & 0.698 \\
  & ViT-B/16   & 28.0 & 19.4 & 23.4 & 0.705 & 0.690 \\
  & ConvNeXt-T & 24.7 & 18.8 & 22.8 & 0.717 & 0.696 \\
\midrule
\multirow[c]{5}{*}{NL-26}
  & AlexNet    & 20.0 & 17.3 & 19.2 & 0.693 & 0.680 \\
  & ResNet-18  & 22.0 & 17.4 & 20.3 & 0.747 & 0.747 \\
  & ResNet-50  & 22.1 & 17.1 & 21.6 & 0.740 & 0.738 \\
  & ViT-B/16   & 24.5 & 17.5 & 18.5 & 0.650 & 0.656 \\
  & ConvNeXt-T & 24.8 & 18.1 & 21.1 & 0.730 & 0.730 \\
\midrule
\multirow[c]{5}{*}{iNat-25}
  & AlexNet    & 5.1 & 3.3 & 5.1 & 0.523 & 0.519 \\
  & ResNet-18  & 8.5 & 6.7 & 7.3 & 0.563 & 0.558 \\
  & ResNet-50  & 9.0 & 6.9 & 7.7 & 0.563 & 0.563 \\
  & ViT-B/16   & 7.2 & 4.7 & 5.6 & 0.539 & 0.541 \\
  & ConvNeXt-T & 7.7 & 5.3 & 6.4 & 0.545 & 0.549 \\
\bottomrule
\end{tabular}
\end{table}

\appsec{app:oracle}{Is the gap repairable by a single temperature?}
The standard fix for miscalibration under shift, re-weighting the seen calibration set
toward the unseen distribution, cannot help here: it borrows the corrective scale from
calibration examples that resemble the test input, and an unseen subtype has almost none.
As a diagnostic rather than a deployable remedy, we therefore fit an
\emph{unseen-oracle} temperature $T^*_{\mathcal U}$ directly on the unseen split, using
its labels and hence unattainable in practice. It is far larger than the seen-fit
$T^*_{\mathcal S}$ ($T^*_{\mathcal U}\approx 2$--$3$ vs.\ $1.2$--$1.7$) and brings the
unseen ECE down to $2$--$5$, so the aggregate overconfidence is a coherent single-scale
error that one temperature could repair in principle. Two caveats keep this from being a
remedy. First, temperature is monotone, so even the oracle leaves the per-sample
discrimination loss of RQ1 untouched: it repairs the average confidence, not the broken
ranking of correct against incorrect predictions. Second, no single temperature serves
both splits at once, since the unseen-oracle scale over-softens the seen inputs (their
ECE rises to ${\approx}10$). It would have to be applied selectively, which requires
knowing which inputs are unseen, and that is the detection question of RQ3b.

\appsec{app:percat}{Per-category heterogeneity.}
The collapse is highly uneven across coarse classes. Table~\ref{tab:app-percat} gives
per-supertype seen/unseen accuracy and ECE (IN-25, ResNet-50), sorted by accuracy
drop: supertypes with few seen subtypes collapse almost completely (\textit{big cat},
accuracy $96.0\%$ seen vs.\ $5.3\%$ unseen, ECE $2$ vs.\ $81$), while richly-sampled ones stay
calibrated (\textit{dog}, 117 subtypes, ECE $1.4$ vs.\ $2.2$), mirroring the $n_j$ trend
of Figure~\ref{fig:imagenet-subtype}.

\begin{table}[htbp]
\centering\footnotesize
\setlength{\tabcolsep}{4pt}
\caption{Per-supertype collapse (IN-25, ResNet-50), sorted by accuracy drop.
$n_j$ = number of seen subtypes; accuracy (\%) and ECE on seen vs.\ unseen.}
\label{tab:app-percat}
\begin{tabular}{lrcccc}
\toprule
Superclass & $n_j$ & Acc$_{\mathcal S}$ & Acc$_{\mathcal U}$ & ECE$_{\mathcal S}$ & ECE$_{\mathcal U}$ \\
\midrule
big cat            & 3   & 96.0 & 5.3  & 2.1 & 80.6 \\
bear               & 2   & 88.3 & 19.3 & 7.6 & 60.0 \\
beverage           & 1   & 92.0 & 32.7 & 4.8 & 33.2 \\
fungus             & 1   & 70.9 & 17.3 & 17.2 & 56.3 \\
aquatic mammal     & 3   & 90.0 & 43.0 & 5.2 & 34.5 \\
utensil            & 5   & 80.7 & 34.7 & 8.4 & 39.2 \\
vegetable          & 11  & 93.3 & 49.7 & 2.9 & 27.5 \\
primate            & 12  & 94.7 & 62.0 & 2.2 & 22.1 \\
building           & 7   & 87.3 & 55.7 & 7.5 & 20.1 \\
insect             & 9   & 97.3 & 66.3 & 1.3 & 17.1 \\
appliance          & 9   & 81.7 & 52.3 & 6.6 & 24.0 \\
footwear           & 4   & 86.7 & 59.7 & 6.1 & 20.6 \\
ungulate           & 11  & 89.3 & 64.3 & 5.9 & 19.8 \\
electronic equip.  & 9   & 90.7 & 67.3 & 4.5 & 13.3 \\
cat                & 5   & 89.7 & 67.0 & 5.8 & 20.2 \\
geological form.   & 5   & 88.7 & 67.7 & 3.0 & 13.4 \\
fruit              & 13  & 92.7 & 71.7 & 2.2 & 11.4 \\
musical instr.     & 15  & 90.3 & 73.0 & 4.0 & 10.9 \\
fish               & 11  & 94.3 & 81.3 & 1.4 & 7.0 \\
clothing           & 33  & 93.0 & 82.0 & 3.2 & 4.9 \\
amphibian          & 8   & 92.3 & 82.0 & 4.5 & 9.8 \\
reptile            & 28  & 93.3 & 85.0 & 3.4 & 5.7 \\
bird               & 34  & 97.7 & 91.3 & 1.5 & 3.5 \\
wheeled vehicle    & 27  & 95.7 & 90.0 & 3.1 & 5.1 \\
dog                & 117 & 98.7 & 96.0 & 1.4 & 2.2 \\
\bottomrule
\end{tabular}
\end{table}

\appsec{app:ksweep}{Fine-label diversity ($k$-sweep).}
Table~\ref{tab:app-ksweep} varies the number of seen fine subtypes per supertype, $k$, on IN-25
(ResNet-50, from scratch) while holding the per-class image budget fixed, so only fine-label diversity
changes. The unseen calibration gap falls monotonically as $k$ grows ($\Delta_{\mathrm{cal}}$ $31\!\to\!23$
from $k{=}1$ to $4$). This upgrades the correlational $n_j$ trend of the Analysis to a causal statement:
training on more fine subtypes causes less overconfidence on unseen ones, holding image count fixed.
(Subtypes are capped here, so the absolute gap is larger than the full-data main-table row.)

\begin{table}[htbp]
\centering\small
\setlength{\tabcolsep}{6pt}
\caption{Fine-label diversity $k$-sweep (IN-25, ResNet-50 from scratch; per-class image budget fixed).
Raising the number of seen subtypes per supertype monotonically shrinks the unseen calibration gap.}
\label{tab:app-ksweep}
\begin{tabular}{cccccc}
\toprule
$k$ & $\mathrm{ECE}_{\mathcal S}$ & $\mathrm{ECE}_{\mathcal U}$ & $\Delta_{\mathrm{cal}}\downarrow$ & Seen acc & Unseen acc \\
\midrule
1 & 1.9 & 32.5 & 30.6 & 79.8 & 26.1 \\
2 & 3.3 & 32.2 & 28.9 & 85.2 & 35.9 \\
3 & 4.0 & 27.8 & 23.8 & 85.6 & 44.4 \\
4 & 3.2 & 25.8 & 22.6 & 85.8 & 45.8 \\
\bottomrule
\end{tabular}
\end{table}

\end{document}